\documentclass[conference]{IEEEtran}
\usepackage{xcolor,soul,framed} 

\colorlet{shadecolor}{yellow}
\usepackage[pdftex]{graphicx}
\graphicspath{{../pdf/}{../jpeg/}}
\DeclareGraphicsExtensions{.pdf,.jpeg,.png}

\usepackage[cmex10]{amsmath}
\usepackage{cite}
\usepackage{rotating}
\usepackage{hyperref} 
\usepackage{tablefootnote}
\usepackage{array}
\usepackage{mdwmath}
\usepackage{amsmath}
\usepackage{amsfonts}
\usepackage{amssymb}
\usepackage{mdwtab}
\usepackage{eqparbox}
\usepackage{url}
\usepackage{xcolor}
\usepackage{comment}
\usepackage{footnote}
\usepackage{multirow}
\usepackage{graphicx}
\usepackage{caption}
\usepackage{algorithm, algorithmicx}
\usepackage{algpseudocode}

\usepackage{float}
\usepackage{placeins}

\usepackage{url} 
\usepackage{array, makecell}
\usepackage{booktabs}
\usepackage{tikz}

\usepackage{algpseudocode}
\ifCLASSINFOpdf
  
\else
 
\fi

\begin{document}
\bstctlcite{IEEEexample:BSTcontrol}
%
\title{A Semantic-Loss Function Modeling Framework With Task-Oriented Machine Learning Perspectives  \\
}

    \author{\IEEEauthorblockN{Ti Ti Nguyen, Thanh-Dung Le, Vu Nguyen Ha, Hong-fu Chou, Geoffrey Eappen, Duc-Dung Tran, \\
    Hung Nguyen-Kha, Prabhu Thiruvasagam, Luis M. Garces-Socarras, Jorge L. Gonzalez-Rios,\\
    Juan C. Merlano-Duncan, Symeon Chatzinotas 
    }\\
    \vspace{-2mm}
    \IEEEauthorblockA{\textit{Interdisciplinary Centre for Security, Reliability and Trust (SnT), University of Luxembourg, Luxembourg} 
	}
    }

\maketitle

\begin{abstract}

The integration of machine learning (ML) has significantly enhanced the capabilities of Earth Observation (EO) systems by enabling the extraction of actionable insights from complex datasets. However, the performance of data-driven EO applications is heavily influenced by the data collection and transmission processes, where limited satellite bandwidth and latency constraints can hinder the full transmission of original data to the receivers.
To address this issue, adopting the concepts of Semantic Communication (SC) offers a promising solution by prioritizing the transmission of essential data semantics over raw information. Implementing SC for EO systems requires a thorough understanding of the impact of data processing and communication channel conditions on semantic loss at the processing center.
This work proposes a novel data-fitting framework to empirically model the semantic loss using real-world EO datasets and domain-specific insights. 
The framework quantifies two primary types of semantic loss: (1) source coding loss, assessed via a data quality indicator measuring the impact of processing on raw source data, and (2) transmission loss, evaluated by comparing practical transmission performance against the Shannon limit. 
Semantic losses are estimated by evaluating the accuracy of EO applications using four task-oriented ML models, \textbf{EfficientViT}, \textbf{MobileViT}, \textbf{ResNet50-DINO}, and \textbf{ResNet8-KD}, on lossy image datasets under varying channel conditions and compression ratios. 
These results underpin a framework for efficient semantic-loss modeling in bandwidth-constrained EO scenarios, enabling more reliable and effective operations.
\end{abstract}


\IEEEpeerreviewmaketitle

\section{Introduction}
\label{sec:Intro}

Earth Observation (EO) utilizes satellite technologies to monitor and analyze Earth's systems, addressing critical global challenges such as climate change, disaster management, agriculture, urban planning, and ecosystem preservation \cite{leyva2023satellite}. Advancements in machine learning (ML) have enhanced EO by enabling efficient analysis of complex datasets, improving accuracy, anomaly detection, and predictive insights \cite{singh2023deep,fontanesi2023artificial}.
For instance, ML algorithms can be used for target detection (e.g., identifying vehicles, ships, or infrastructure), target tracking (e.g., monitoring the movement of deforestation or urban sprawl), segmentation (e.g., delineating land use types or vegetation classes), and level estimation (e.g., estimating crop yield or water usage at regional levels) \cite{affek2024survey}.
However, realizing EO applications requires seamless and efficient data exchange between satellites and ground stations. Satellites capture enormous amounts of raw data, which often need to be transmitted to mobile edge computing (MEC) satellites or ground stations equipped with high-power AI processors for enabling rapid, scalable, and actionable intelligence for various applications \cite{chou2023edge}. The transmission of this data in satellite systems faces significant challenges due to bandwidth constraints and latency requirements \cite{mahboob2024revolutionizing}.

Semantic communication (SC) has emerged as a transformative paradigm, shifting the focus from traditional bit-oriented communication to transmitting the meaning or intent behind the data. Unlike conventional methods, which often transmit large volumes of raw and redundant information, SC prioritizes the communication of essential semantics \cite{yang2022semantic,chou2024air}. This approach enhances EO applications by reducing communication overhead and latency in bandwidth-limited scenarios, but challenges persist due to the lack of models linking EO objectives to transmitted data \cite{qin2021semantic,chou2024cognitive}. Unlike traditional communication systems, where performance can be optimized based on well-defined metrics like bit error rate or signal-to-noise ratio, SC requires the transmission of data that directly aligns with the meaning or intent of the application. Without a clear model, it becomes difficult to quantify what constitutes `meaningful' or `sufficient' data for a specific EO task. This uncertainty complicates the design of SC systems, as they must dynamically learn and adapt to prioritize and interpret data relevance in real-time, often relying on machine learning models or heuristic methods \cite{liu2023EfficientViT, mehta2022separable, le2024semantic, goldblum2024battle}. 
Moreover, the stochastic nature of wireless channels and limited bandwidth exacerbate these challenges, necessitating innovative solutions to bridge the gap between raw data exchange and application-driven semantic objectives.
To address these challenges, models are required to represent the relationship between EO objectives, communication conditions, and the data. 
For example, in agricultural monitoring, with a lower SNR, SC may prioritize sending lower-resolution data, which could reduce the overall accuracy but still provide enough information for rough yield estimates. In urban planning, where long-term trends are more important than real-time data, SC can tolerate higher source loss and lower accuracy, transmitting lower-quality images to assess urban sprawl over time.

This work introduces a practical data-fitting approach to model the SC process for EO applications. Using real-world datasets and application-specific insights, this framework empirically captures the relationship between EO objectives and the transmitted and received data. Specifically, EuroSAT \cite{helber2019eurosat} dataset is utilized for land use and land cover classification.  Effective ML models,  including \textbf{EfficientViT} \cite{liu2023EfficientViT}, \textbf{MobileViT} \cite{mehta2022separable}, \textbf{ResNet8-KD} (trained via knowledge distillation) \cite{le2024semantic} and \textbf{ResNet50-DINO} (pre-trained by Facebook) \cite{goldblum2024battle}, are implemented to capture task accuracy as the EO objective. These models are deployed in  ground stations to utilize high-power AI capacity. To reduce the amount of transmitted data, raw data is processed at various quality levels. Additionally, the framework incorporates transmission loss, defined as the gap between practical transmission capabilities and the theoretical Shannon limit, into its evaluation of EO task accuracy. These losses are integrated into task-oriented semantic loss modeling, represented mathematically as the sum of the product of a shifted sigmoid function and an exponential function. By addressing these losses, the proposed framework provides a comprehensive foundation for modeling SC in bandwidth-constrained EO scenarios, paving the way for more efficient and reliable EO operations.  



\section{Data Acquisition for Semantic-Loss Modeling}

\subsubsection{Original Dataset}
The public benchmark for land use and land cover classification dataset, EuroSAT \cite{helber2019eurosat}, will be used in this study. This is a large-scale benchmark dataset designed explicitly for land use and land cover classification derived from Sentinel-2 satellite imagery. It comprises 27,000 geo-referenced labeled images, each measuring 64x64 pixels and spanning 13 spectral bands. The dataset is categorized into 10 classes. Each
class contains 2000–3000 images, including industrial buildings, residential buildings, annual crops, permanent crops, rivers, seas and lakes, Herbaceous vegetation, highways, pastures, and forests across Europe. Due to its compact image size and diverse class representation, EuroSAT is particularly well-suited for developing and evaluating DL models intended for onboard satellite processing in EO missions. This makes it a valuable resource for applications such as real-time environmental monitoring, disaster response, and precision agriculture, where in-situ processing capabilities are crucial for timely decision-making.

\subsubsection{Satellite communication-integrated dataset acquisition}
\label{sec_2b}
\begin{figure}
    \centering
    \includegraphics[width=0.8\linewidth]{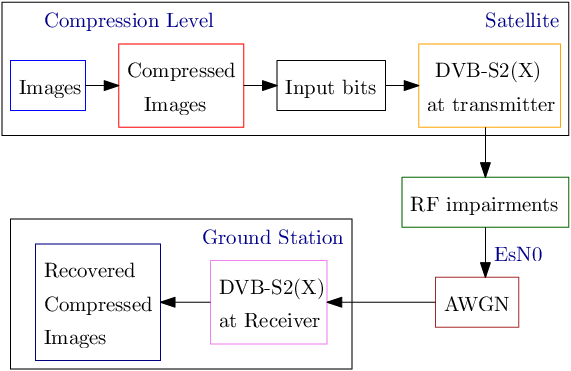}
    \captionsetup{font=small}
    \caption{Block diagram of SatCom-integrated data acquisition}
    \label{fig:blockdiagram}
    \vspace{-5mm}
\end{figure}

In this paper, we explore the application of ML models for image classification at ground stations. The block diagram of satellite communication (SatCom)-integrated data acquisition is shown in Fig.~\ref{fig:blockdiagram}. Specifically, images captured by EO satellites (i.e., original dataset) are transmitted to ground stations for processing on colocated servers. The images are assumed to be transmitted using the DVB-S2(X) standard, a well-established protocol for SatComs \cite{etsi_dvbs2x,ha2022geo}. 
Our communication system model accounts for imperfections in the RF front-end, including carrier frequency offset (CFO), sampling clock offset (SCO), and phase noise at the receiver. The image data stream, affected by these impairments, is transmitted through an additive white Gaussian noise (AWGN) channel.
To mitigate transmission errors, the DVB-S2(X) standard employs Bose–Chaudhuri–Hocquenghem (BCH) and low-density parity check (LDPC) codes as part of its forward error correction (FEC) mechanisms. At the receiver, operations such as matched filtering, timing recovery, and carrier recovery are performed to reconstruct the transmitted data \cite{mathworks_dvbs2_simulation}. 
The DVB-S2(X) parameters are configured in a Matlab software simulator, with a CFO of 3 MHz and a SCO of 5. The system operates under a low phase noise environment, with phase noise levels of -73 dBc/Hz at 100 Hz, -83 dBc/Hz at 1 kHz, -93 dBc/Hz at 10 kHz, -112 dBc/Hz at 100 kHz, and -128 dBc/Hz at 1 MHz \cite{mathworks_dvbs2_simulation}. 

On the other hand, to reduce the amount of transmitted data, lossy image compression is applied. Specifically, Matlab’s `imwrite' function is used to generate compressed images at quality levels of 10, 20, 30, 40, 50, 60, 70, 80, 90, and 100. Additionally, in SC, when precise bit transmission is not the primary criterion, the data may experience further losses due to wireless transmission constraints. This can occur when more data is transmitted than the channel's capacity allows to meet the delay requirement, even if it results in a higher bit error rate compared to traditional communication. 
To quantify this, we introduce the actual to Shannon-based signal-to-noise-plus-interference ratio (SNR) ratio $s$, which is defined as
\begin{equation}
    s = \frac{\gamma}{\gamma_{\sf Shannon}},
\end{equation}
where $\gamma$ is the received SNR (in dB) and $\gamma_{\sf Shannon}$ is the corresponding SNR to the actual transmission rate per Hz $r$, which is given by
$   \gamma_{\sf Shannon} = 10\log_{10} \left( 2^r-1\right)$.

\begin{figure*}[t] 
    \centering
    \includegraphics[width=0.95\textwidth]{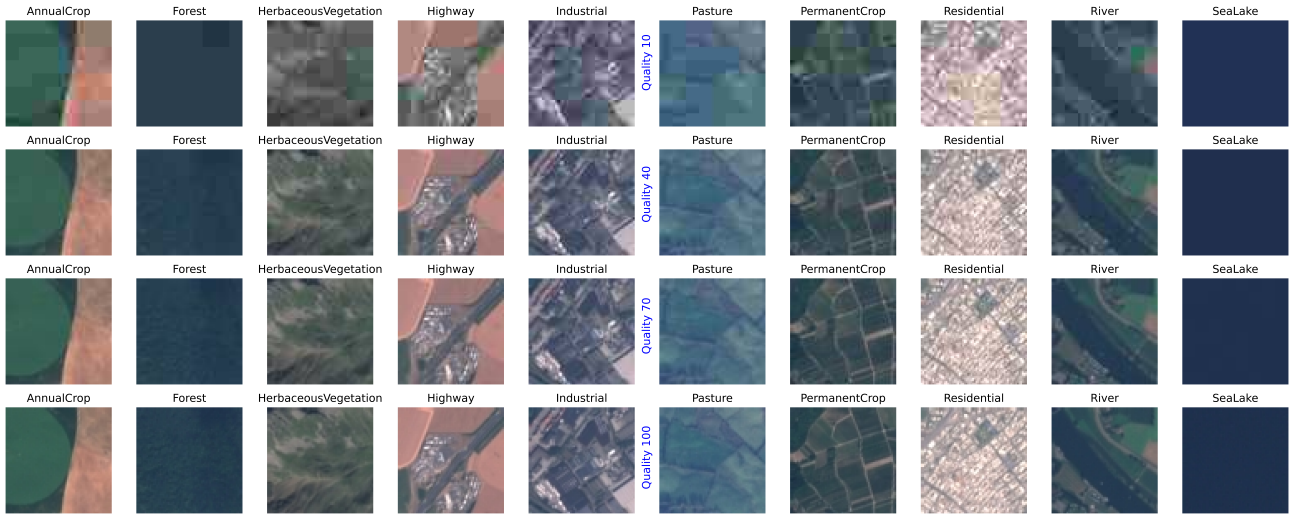}
    \captionsetup{font=small}
    \caption{The loss resulting from source information reduction.}
    \label{fig:loss_quality}
    \vspace{-3mm}
\end{figure*}

\begin{figure*}[h] %
    \centering
    \includegraphics[width=0.95\textwidth]{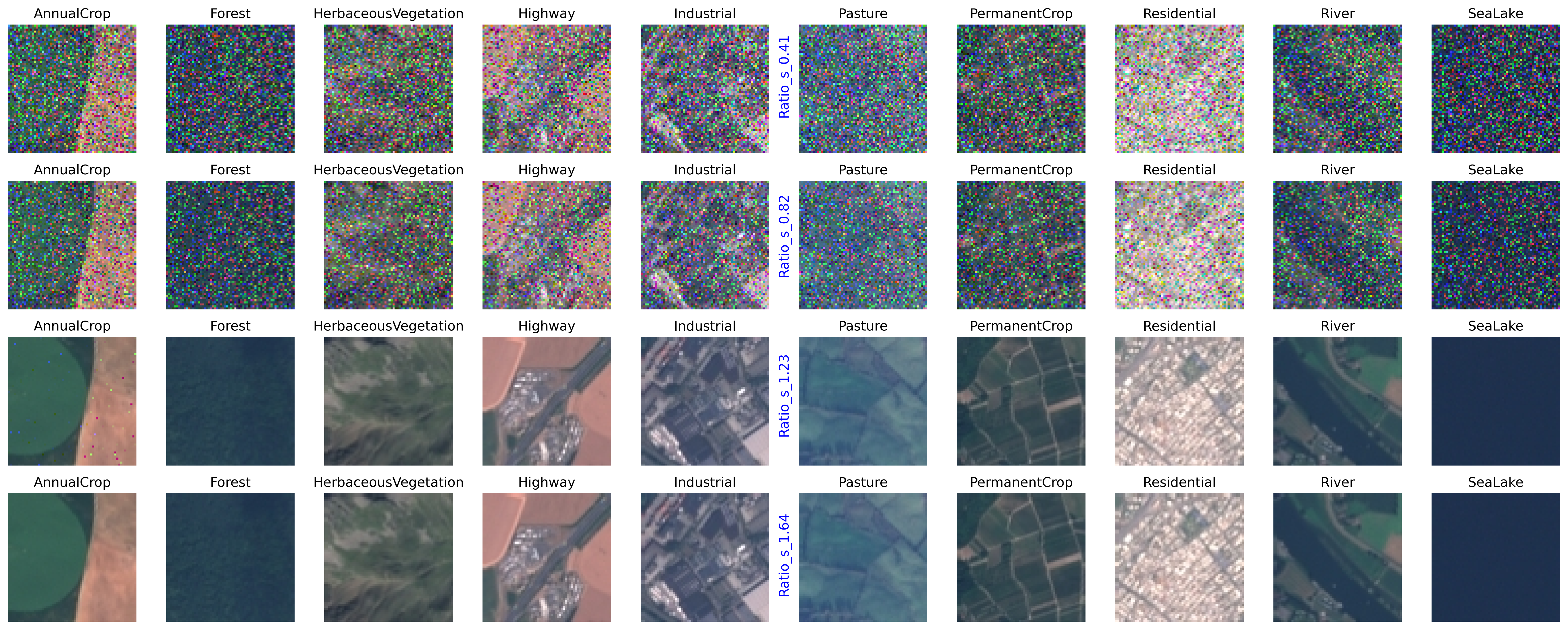}
    \captionsetup{font=small}
    \caption{The loss resulting from imperfect wireless transmission.}
    \label{fig:loss_SNR}
    \vspace{-3mm}
\end{figure*}

Depending on the application’s latency sensitivity and the rate-limited channel conditions, ratio $s$ can take on different values. In this paper, four levels of ratio $s$ are considered: 0.41, 0.82, 1.23, and 1.64. For each combination of compression quality $q$ and  ratio $s$, a new recovered dataset is generated using the DVB-S2(X) standard. Starting from the original EuroSAT dataset, we create  40 new datasets, corresponding to $10$ compression quality levels $q$ multiplied by $4$  levels of ratio $s$.
Figures~\ref{fig:loss_quality} and \ref{fig:loss_SNR} illustrate the impact of source reduction loss and transmission loss on the received data, respectively. Specifically, we randomly select one image from each class of the EuroSAT dataset and present their corresponding results for each value of $q$ and $s$.

\section{Methodology - Curve Fitting-based Modeling}
\vspace{-0.1 cm}
A recent study \cite{le2024board} provides a comprehensive analysis of Vision Transformers (ViTs) in EO Image Classification (EO-IC), identifying EfficientViT \cite{liu2023EfficientViT} as one of the most effective models based on its outstanding balance of performance, robustness, and computational efficiency. These findings highlight  the suitability of EfficientViT for enabling EO applications effectively. 
From a technical standpoint, \textbf{EfficientViT} employs a hybrid architecture that combines convolutional layers with local window attention mechanisms. This design enables high classification accuracy while optimizing computational efficiency, making it particularly suitable for latency-sensitive applications \cite{liu2023EfficientViT}. 
In this paper, we select the EfficientViT for evaluating the EO application. 
After applying the EfficientViT algorithm to the datasets described in Section~\ref{sec_2b}, we obtain the results as shown in Table~\ref{tab:results_EfficientViT}.

\begin{table}[!t]
    \centering
    \caption{\small\textsc{Empirical accuracy $\boldsymbol{Y}$ based on EfficientViT.}}
    \label{tab:results_EfficientViT}
    \begin{tabular}{@{}c*{4}{>{\centering\arraybackslash}p{1cm}}@{}}
        \toprule
        \( q \backslash s \) & 0.41 & 0.82 & 1.23 & 1.64 \\ 
        \midrule
        10  & 81.94 & 81.94 & 88.42 & 88.50 \\ 
        20  & 83.92 & 86.75 & 92.91 & 93.72 \\ 
        30  & 84.95 & 87.57 & 95.02 & 95.51 \\ 
        40  & 85.67 & 87.77 & 95.92 & 96.31 \\ 
        50  & 85.99 & 88.38 & 96.33 & 96.74 \\ 
        60  & 86.17 & 88.54 & 96.65 & 96.95 \\ 
        70  & 86.23 & 88.62 & 96.87 & 97.31 \\ 
        80  & 86.28 & 88.87 & 97.07 & 97.58 \\ 
        90  & 86.32 & 89.17 & 97.64 & 97.94 \\ 
        100 & 86.39 & 89.37 & 98.05 & 98.37 \\ 
        \bottomrule
    \end{tabular}
    \vspace{-4mm}
\end{table}

To the best of our knowledge, there is no theoretical result in the literature for evaluating the EO application objective, particularly in the context of losses arising from source information reduction and wireless transmission. To address this gap and gain insights into modeling the SC, we first executed EO applications under varying control parameter settings for $q$ and $s$. The results, summarized in Table~\ref{tab:results_EfficientViT}, use accuracy as the primary metric for assessing the EO application objective.

\begin{figure}[!t]
    \centering
\includegraphics[width=1\linewidth]{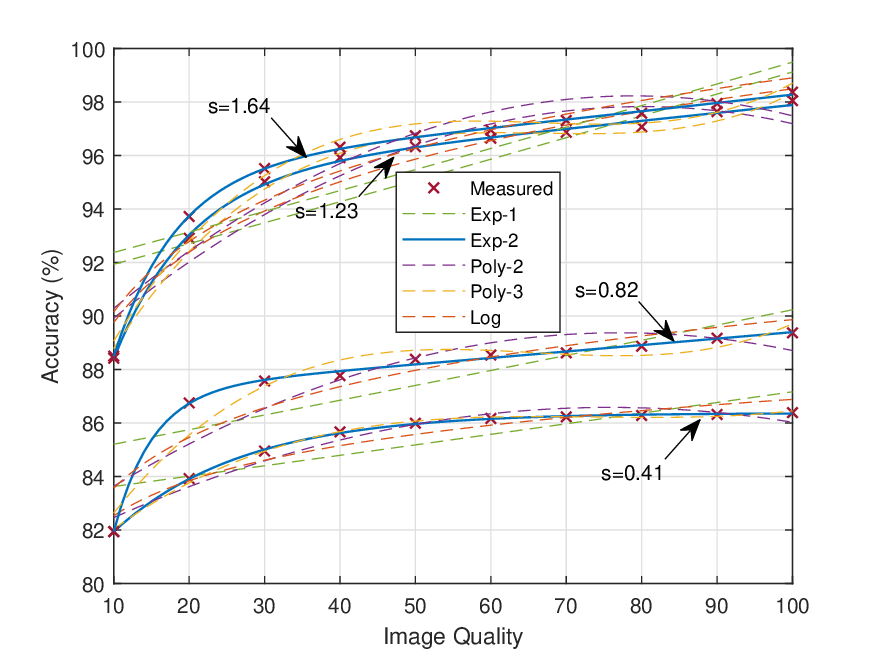}
    \captionsetup{font=small}
    \caption{Curve fitting model with different models.}
    \label{fig:1Dq}
    \vspace{-3mm}
\end{figure}

In the absence of theoretical models, we employ a practical data-fitting approach to represent the EO application objective as a non-linear function of $q$ and $s$. Generally, choosing data-fitting models to optimize system parameters can impact system performance \cite{nguyen2019joint}. A more accurate fitting model typically leads to a more reliable and effective design.
The fitting process begins with one-dimensional fitting, where we derived closed-form relationships for accuracy as a function of $q$ (for a fixed $s$). The results are also analyzed with $s$. These initial models are served as a foundation for developing a general fitting model that captures the combined influence of $q$ and $s$ on the EO application objective.

\subsubsection{1-dimensional fitting model}
For given $s$, we use the MATLAB curve fitting tool and consider several popular one-dimensional fitting models, including `Poly-2', `Poly-3', `Log', `Exp-1', `Exp-2', which are defined as follows:
\begin{equation}
    \begin{aligned}
        \text{Poly-2}(x) &= ax^2 + bx +c, \\
        \text{Poly-3}(x) &= ax^3 + bx^2 +cx+d, \\
        \text{Log}(x) &= a\log(x)+b,\\
        \text{Exp-1}(x) &= a\exp(bx),\\
        \text{Exp-2}(x) &= a\exp(bx) + c\exp(dx).
    \end{aligned}
    \label{eq_fit1}
\end{equation}

The experimental data in Table~\ref{tab:results_EfficientViT} are used to fit the parameters of the functions shown in equation \eqref{eq_fit1}. As illustrated in Fig.~\ref{fig:1Dq}, `Exp-2' is the best model for expressing accuracy as a function of $q$ for a given $s$. 

\begin{figure}[!t]
    \centering
    \includegraphics[width=0.8\linewidth, height=45mm]{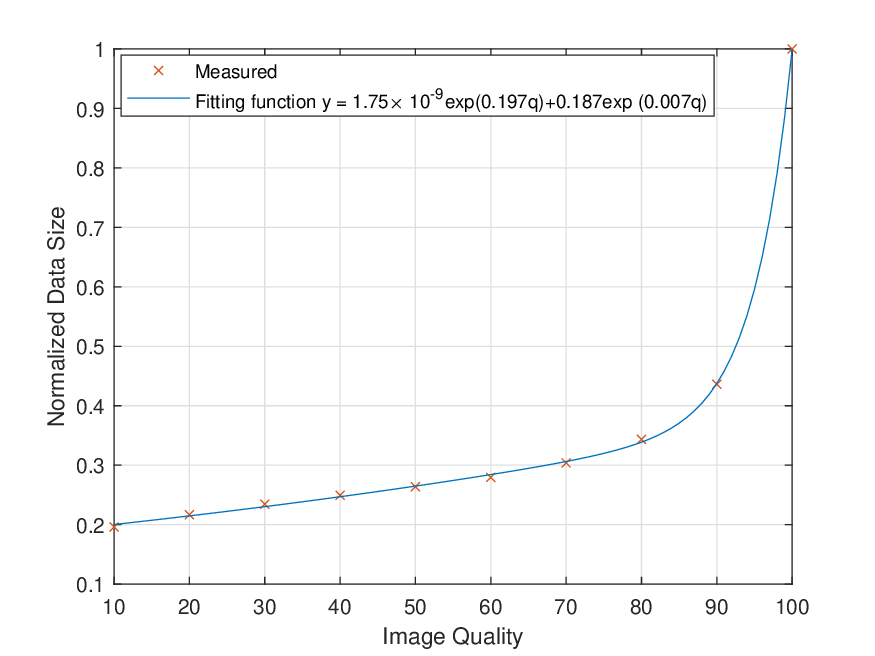}
    \captionsetup{font=small}
    \caption{Normalized data size using `Exp-2' fitting model.}
    \label{fig:datasize}
    \vspace{-3mm}
\end{figure}

Furthermore, the `Exp-2' model can also be used to capture the relationship between the normalized data size and image quality $q$, as shown in Fig.~\ref{fig:datasize}. Note that the normalized data size is defined as the ratio of the data size corresponding to quality $q$ to the data size at a quality of 100.

\begin{table}[!t]
    \centering
    \caption{\small\textsc{Parameters corresponding to `Exp-2' model.}}
    \label{tab:results_fit1}
    \begin{tabular}{@{}c*{4}{>{\centering\arraybackslash}p{1.25cm}}@{}}
        \toprule
        \( s \backslash \text{Exp-2 Param.}  \) & a & b & c & d \\ 
        \midrule
        0.80  & 86.4433 & -7.624e-06 & -8.0206 & -0.0578 \\ 
        0.94  & 87.0050 & 2.711e-04 & -37.5884 & -0.1959 \\ 
        1.08  & 94.9856 & 3.011e-04 & -18.7067 & -0.1001 \\ 
        1.24  & 95.2171 & 3.160e-04 & -23.7064 & -0.1217 \\
        \bottomrule
    \end{tabular}
    \vspace{-3mm}
\end{table}

\begin{figure}[!t]
    \centering
    \includegraphics[width=0.8\linewidth, height=45mm]{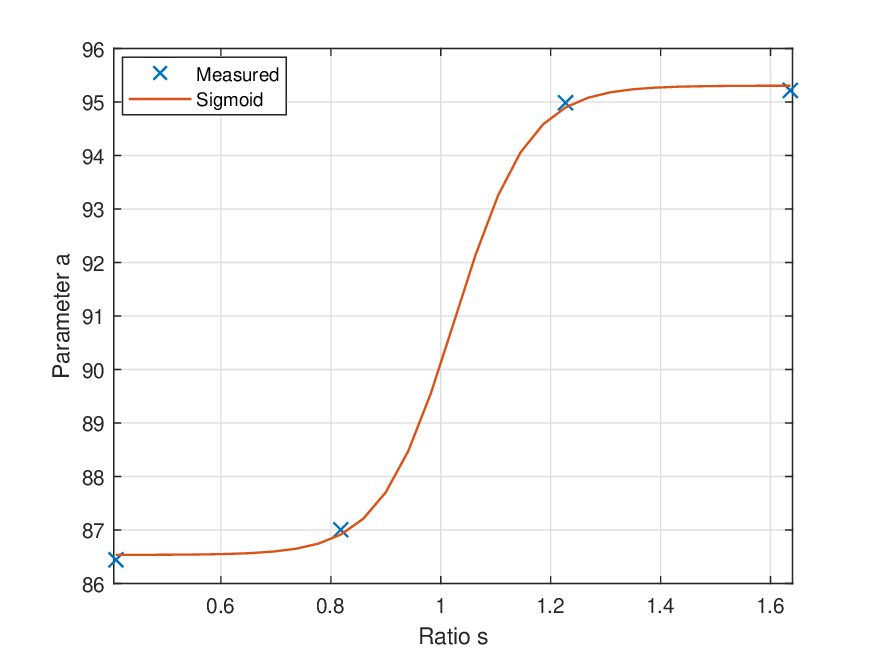}
    \captionsetup{font=small}
    \caption{Curve fitting model for parameter $a$ of `Exp-2' in Table~\ref{tab:results_fit1}.}
    \label{fig:1Ds}
    \vspace{-0.5 cm}
\end{figure}

When `Exp-2' model is considered to fit the data in Table~\ref{tab:results_EfficientViT}, their corresponding parameters $a, b, c$ and $d$ in equation \eqref{eq_fit1} are provided in Table~\ref{tab:results_fit1}. On the other hand, as shown in Fig.~\ref{fig:1Ds}, parameter $a$ in Table~\ref{tab:results_fit1} can be fitted  using the shifted sigmoid functions, as follows
\begin{equation}
    \text{Sigmoid}(x) = b +  \frac{c}{1 + \exp(-dx-e)},
    \label{eq_sig}
\end{equation}
where $b = 95.3055$, $c = -8.7716$, $d = -14.9563$, and $e= 15.3302$.

\subsubsection{A general fitting model}
Based on the fitting models in \eqref{eq_fit1} and  \eqref{eq_sig}, we propose a general fitting model for accuracy concerning both parameters $q$ and $s$ as follows
\vspace{-3mm}
\begin{equation}
    \xi_{\sf acc} = \mu_{0} + \sum_{i=1}^{N_{\sf c}} \left( \mu_{1,i} + \frac{\mu_{2,i}}{1 + \exp \left(-\mu_{3,i} s - \mu_{4,i}\right)} \right) \exp(\mu_{5,i} q),
    \label{eq_2dfit}
\end{equation}
where $N_{\sf c}$ is the number of terms.

\begin{figure}[!t]
    \centering
\includegraphics[width=0.9\linewidth]{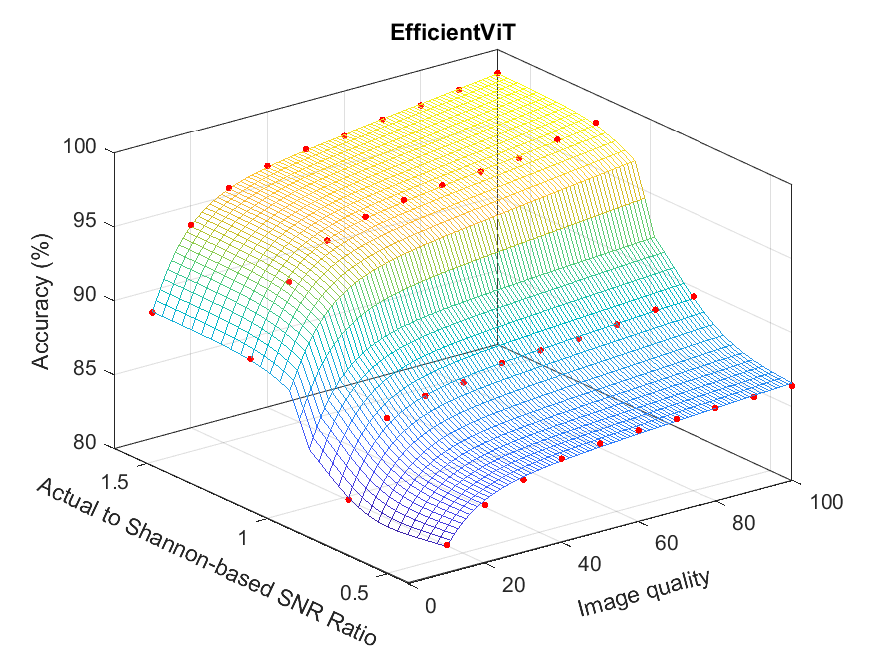}
    \captionsetup{font=small}
    \caption{Curve fitting model for the EfficientViT case ($N_{\sf c} = 4$).}
    \label{fig:enter-label1}
    \vspace{-3mm}
\end{figure}

However, the MATLAB curve fitting tool does not support the multi-dimensional fitting with the structure given in  \eqref{eq_2dfit}. To find the parameters $\mu_0, \mu_{i,j},  i=1:5, j=1:N_{\sf c}$, we employ the gradient descent optimization method. The details of this approach are provided in Algorithm~\ref{alg1}, where lines 4-5 to determine the fitting values $\boldsymbol{\xi}_{\sf acc}$ and the error between the fitted results $\boldsymbol{\xi}_{\sf acc}$ and the measured data $\boldsymbol{Y}$. Lines $6-23$ are to compute the gradients of $\xi_{\sf acc}$ with respect to $\boldsymbol{\mu}_i, i=0:5$, while line $24$ is to update $\boldsymbol{\mu}_i, i=0:5$. In this paper, we set $\alpha_0 = \alpha_1 = \alpha_2 = 1e-3$, $\alpha_3 = \alpha_4 = 1e-4$, and $\alpha_5=1e-11$.
Fig.~\ref{fig:enter-label1} shows the fitting model \eqref{eq_2dfit}  when the Algorithm~\ref{alg1} with $N_{\sf c} = 4$ converges. The fitting model parameters corresponding to the results illustrated in Fig.~\ref{fig:enter-label1} are given in Table~\ref{tab:results_fig7}.

\begin{algorithm}[!t]
\caption{\small\textsc{Gradient Descent for SC Modeling}}\label{alg1}
\small
\begin{algorithmic}[1]
\State \textbf{Initialize:} Set parameters $\boldsymbol{\mu}_{i}$ and learning rates $\alpha_i, i=0:5$, number of terms $N_{\sf c}$,  number of iterations $N$, accuracy data matrix $\boldsymbol{Y}$ and set of control parameters $\boldsymbol{q}$ and $\boldsymbol{s}$.
\State Initialize gradients: ${\sf{Grad}}_{\mu_0} \gets 0$, and ${\sf{Grad}}_ {\mu_i} \gets \boldsymbol{0}$, for $i=1:5$.
\For{$\text{iter} = 1$ to $N$}        
    \State Compute  $\xi_{\sf acc}$ as in \eqref{eq_2dfit}.
    \State Compute error matrix: $\Psi \gets \boldsymbol{Y} - \xi_{\sf acc}$.
    \For{$i = 1$ to $\text{length}(s)$}
        \For{$j = 1$ to $\text{length}(q)$}
            \For{$k = 1$ to $N_{\sf c}$}
                \State Compute $\sigma_k \gets \frac{1}{1 + \exp(-\mu_{3,k} s_i - \mu_{4,k})}$.
                \State Compute $\beta_k \gets \exp(\mu_{5,k} q_j)$.
                \State {Update gradients: (where $\epsilon_{ij} = [\Psi]_{i,j}$)\\
                \hspace{2 cm} ${\sf{Grad}}_{\mu_{1,k}} \gets {\sf{Grad}}_{\mu_{1,k}} - 2 \epsilon_{ij} \beta_k$. \\
                \hspace{2 cm} ${\sf{Grad}}_{\mu_{2,k}} \gets {\sf{Grad}}_{\mu_{2,k}} - 2  \epsilon_{ij} \beta_k \sigma_k$.\\
                \hspace{2 cm} ${\sf{Grad}}_{\mu_{3,k}} \gets {\sf{Grad}}_{\mu_{3,k}} $\\
                \hspace{4 cm} $ - 2  \epsilon_{ij}  \beta_k \mu_{2,k} \sigma_k (1 - \sigma_k) s_i$. \\
                \hspace{2 cm} ${\sf{Grad}}_{\mu_{4,k}} \gets {\sf{Grad}}_{\mu_{4,k}}  $\\  
                \hspace{4 cm}$ - 2 \epsilon_{ij} \beta_k \mu_{2,k} \sigma_k (1 - \sigma_k).$\\
                \hspace{2 cm} ${\sf{Grad}}_{\mu_{5,k}} \gets {\sf{Grad}}_{\mu_{5,k}} $\\  
                \hspace{4 cm}$ - 2  \epsilon_{ij}  \beta_k q_j (\mu_{1,k} + \mu_{2,k}  \sigma_k).$}
            \EndFor
            \State ${\sf{Grad}}_{\mu_0} \gets {\sf{Grad}}_{\mu_0} - 2 \epsilon_{ij}.$
        \EndFor
    \EndFor
    \State Update: $\mu_{0} \gets \mu_{0} - \alpha_0  {\sf{Grad}}_{\mu_0}$
    \State Update: $\mu_{i,k} \gets \mu_{i,k} - \alpha_i  {\sf{Grad}}_{\mu_{i,k}}$, $i=1:5$, $k=1:N_{\sf c}$.    
\EndFor
\State \textbf{Output:} Final parameters $\boldsymbol{\mu}_i$, $i=0:5$.
\end{algorithmic}
\normalsize
\end{algorithm}

\begin{table}[!t]
    \centering
    \caption{\small\textsc{Parameters of the proposed model in Fig.~\ref{fig:enter-label1}.}}
    \label{tab:results_fig7}
    \begin{tabular}{@{}c*{5}{>{\centering\arraybackslash}p{1cm}}@{}}
        \toprule
        \( i \backslash \text{Param.}  \)  & $\mu_{i,1}$ & $\mu_{i,2}$ & $\mu_{i,3}$ & $\mu_{i,4}$ & $\mu_{i,5} \times 1e3$ \\ 
        \midrule
        1  & 151.72 & 8.33 & 8.77 & -7.89 & 0.17 \\ 
        2  & -25.38 & 16.23 & -5.84 & 4.18 & -123.03  \\ 
        3  & -1.44 & 2.08 & 213.56 &-218.38 & 4.96 \\ 
        4  & 155.01 & -118.34 & -68.98 &70.58 & 581.15 \\
        \hline
        $\mu_{0}$ & \multicolumn{5}{c}{-65.51} \\
        \bottomrule
    \end{tabular}
    \vspace{-3mm}
\end{table}






\section{Results and Discussions}
In this section, we demonstrate the fitting capability of the proposed model and compare it with data obtained using other well-known machine learning algorithms, including \textbf{EfficientViT}, \textbf{MobileViT}, \textbf{ResNet50-DINO}, and \textbf{ResNet8-KD}. 
\textbf{MobileViT} leverages a combination of convolutional and transformer-based processing. Through the use of depthwise separable convolutions and self-attention mechanisms, MobileViT achieves high accuracy with minimal computational overhead, further reinforcing its practicality for EO-IC tasks \cite{mehta2022separable}.
We also include two ResNet-based models as part of our baseline: \textbf{ResNet8-KD} (trained via knowledge distillation) \cite{le2024semantic} and \textbf{ResNet50-DINO} (pre-trained by Facebook). ResNet8-KD, a compact variant of ResNet, is effective in resource-constrained scenarios due to its ability to balance computational efficiency with classification performance. This model was trained using knowledge distillation, a process that leverages the predictions of a larger teacher model to improve the performance of a smaller, more efficient student model. 

In contrast, \textbf{ResNet50-DINO} \cite{goldblum2024battle} represents a powerful pre-trained model. Developed by Facebook, it employs self-supervised learning with DINO (Distillation with No Labels), which enables the model to learn rich feature representations from unlabeled data. ResNet50-DINO has demonstrated exceptional performance in various image classification tasks, making it a robust baseline for evaluating image classification in EO. 
The combination of these four models—EfficientViT, MobileViT, ResNet8-KD, and ResNet50-DINO - provides a diverse and comprehensive baseline for assessing performance, computational efficiency, and robustness in EO applications.

\begin{figure}[!t]
    \centering
    \includegraphics[width=0.9\linewidth]{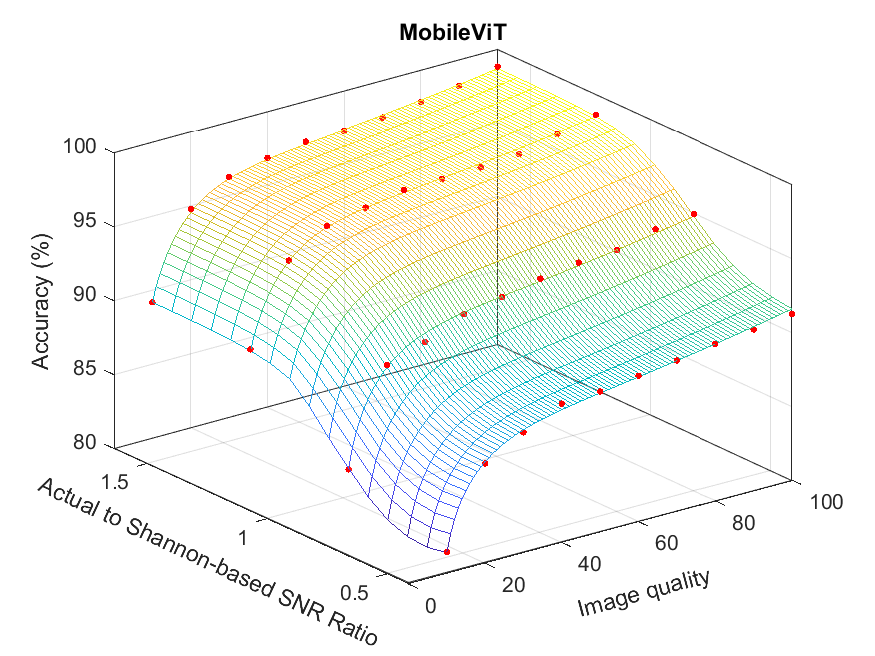}
    \captionsetup{font=small}
    \caption{Curve fitting model for the MobileViT case ($N_{\sf c} = 4$).}
    \label{fig:Mobile_ViT}
    \vspace{-3mm}
\end{figure}

\begin{figure}[!t]
    \centering
    \includegraphics[width=0.9\linewidth]{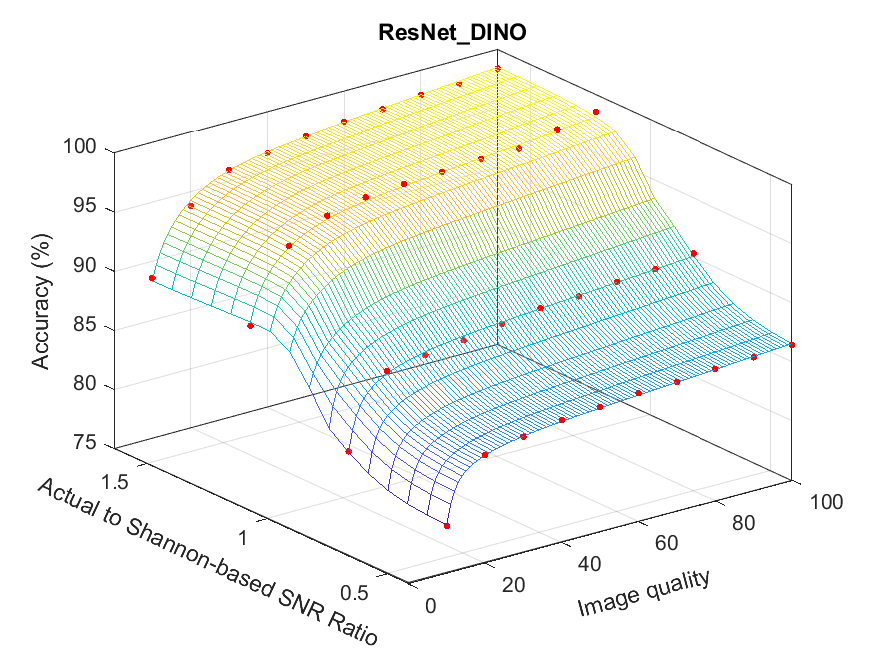}
    \captionsetup{font=small}
    \caption{Curve fitting model for the ResNet\_DINO case ($N_{\sf c} = 4$).}
    \label{fig:DINO}
    \vspace{-3mm}
\end{figure}

\begin{figure}[!t]
    \centering
    \includegraphics[width=0.9\linewidth]{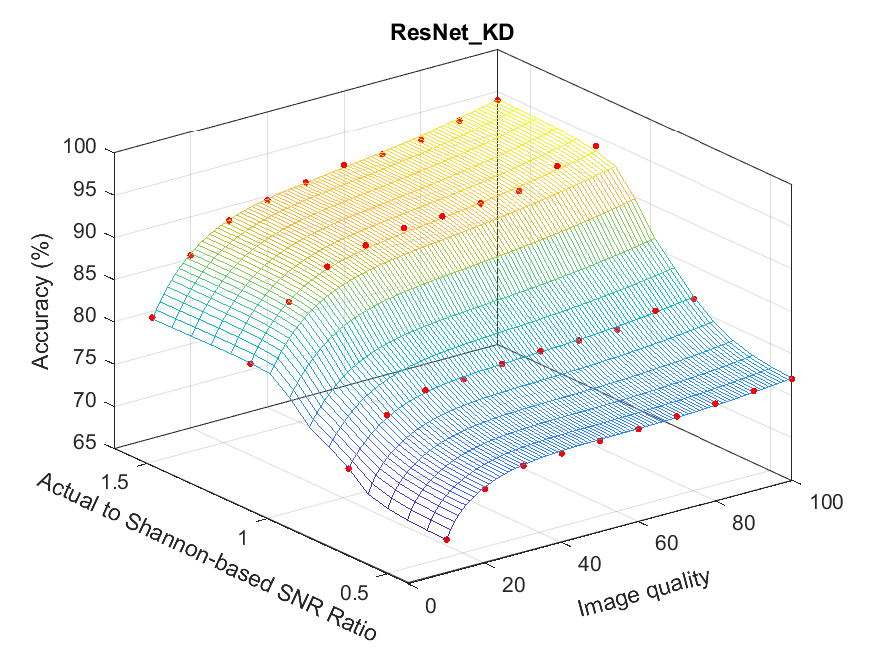}
    \captionsetup{font=small}
    \caption{Curve fitting model for the ResNet\_KD case ($N_{\sf c} = 4$).}
    \label{fig:KD}
    \vspace{-3mm}
\end{figure}

\begin{figure}[!t]
    \centering
    \includegraphics[width=0.8\linewidth]{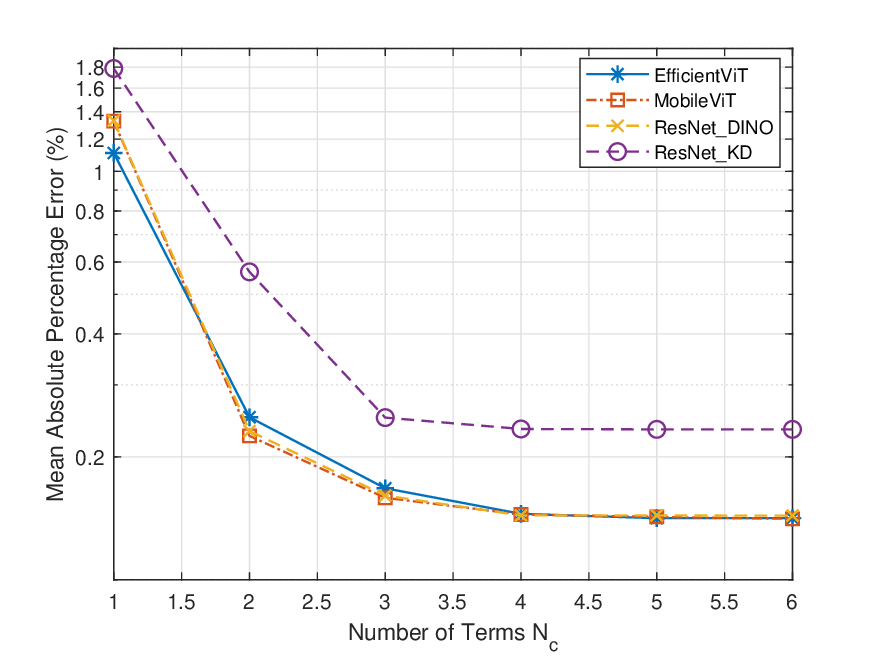}
    \captionsetup{font=small}
    \caption{MAPE vs. number of terms for model fitting.}
    \label{fig:Number_Term}
    \vspace{-3mm}
\end{figure}

Figs.~\ref{fig:Mobile_ViT}, ~\ref{fig:DINO}, and~\ref{fig:KD}  show that the proposed model in \eqref{eq_2dfit} fits well in different scenarios with various ML algorithms. This includes both low-complexity algorithms, such as \textbf{ResNet8-KD}, and higher-complexity algorithms, such as \textbf{EfficientViT}, \textbf{MobileViT}, and \textbf{ResNet50-DINO}. On the other hand, the mean absolute percentage error (MAPE) is used to evaluate how well the proposed model fits the measured data. As shown in Fig.~\ref{fig:Number_Term}, the MAPE value decreases as $N_{\sf c}$  increases, reaching a significantly small value of less than 0.25 \% when the number of terms $N_{\sf c}$ is equal or greater than 4.

\section{Conclusions}
This work presents a novel data-fitting framework for modeling SC in EO applications, specifically showing the relation between the EO objective and the quality of transmitted data as well as the channel transmission condition. By integrating real-world datasets and application-specific insights, the framework empirically captures the relationship between EO objectives and the transmitted data. Through the use of a shifted sigmoid function and exponential function, the framework offers a comprehensive approach for modeling SC in EO scenarios considering source reduction loss and transmission loss. Furthermore, the proposed model is also verified by the most advanced ML techniques, including EfficientViT, MobileViT, ResNet-DINO, and ResNet-KD. Our future work will investigate novel and emerging EO applications to further evaluate the applicability of the proposed model to these use cases. We also plan to explore the application of this proposed SC model to optimize data exchange and resource management, enhancing the real-time performance of EO applications.

\section*{Acknowledgment}
This work was funded by the Luxembourg National Research Fund (FNR), with granted SENTRY project corresponding to grant reference C23/IS/18073708/SENTRY.

\bibliographystyle{IEEEtran}

\end{document}